\pdfoutput=1
\documentclass[aos]{imsart}

\RequirePackage{amsthm,amsmath,amsfonts,amssymb}
\RequirePackage[authoryear]{natbib}
\RequirePackage[colorlinks,citecolor=blue,urlcolor=blue]{hyperref}
\RequirePackage{graphicx}

\startlocaldefs
\makeatletter
\def\journal@name{Accepted for publication in the Annual Review of Statistics and Its Application, Volume 14 }
\def\journal@url{https://doi.org/10.1146/annurev-statistics-043025-102547}
\makeatother
\def\boldq{\boldsymbol q}
\def\boldp{\boldsymbol p}
\def\bfy{\mathbf y}
\endlocaldefs

\begin{document}

\begin{frontmatter}
\title{A Statistical Perspective on Knowledge Distillation: Foundations, Classical Methods, and Large Language Model Extensions\thanksref{T1}}
%% Self-archiving statement required by the Annual Reviews Copyright License (Authors' Rights, item 5):
%%   (a) state by which Annual Reviews journal the manuscript was accepted;
%%   (b) after publication, include the sentence "Posted with permission ... https://www.annualreviews.org." verbatim.
%% The DOI link to the published version is optional (replace with the AR ePrint URL if desired).
\thankstext{T1}{Accepted for publication in the \textit{Annual Review of Statistics and Its Application}, Volume~14;
published version: \url{https://doi.org/10.1146/annurev-statistics-043025-102547}.
Posted with permission from the Annual Review of Statistics and Its Application, Volume 14;
copyright 2027 the author(s), \url{https://www.annualreviews.org}.}
\runtitle{A Statistical Perspective on Knowledge Distillation}
\runauthor{L. Fang et al.}

\begin{aug}
\author[A]{\fnms{Luyang}~\snm{Fang}}
\author[A]{\fnms{Haoran}~\snm{Lu}}
\author[A]{\fnms{Jiazhang}~\snm{Cai}}
\author[A]{\fnms{Tao}~\snm{Wang}}
\author[B]{\fnms{Huimin}~\snm{Cheng}}
\author[A]{\fnms{Wenxuan}~\snm{Zhong}}
\author[A]{\fnms{Ping}~\snm{Ma}\ead[label=e1]{pingma@uga.edu}}

\address[A]{Department of Statistics, University of Georgia, Athens, Georgia 30602, USA\printead[presep={,\ }]{e1}}
\address[B]{Department of Biostatistics, Boston University, Boston, Massachusetts 02118, USA}
\end{aug}

\begin{abstract}
Knowledge Distillation (KD) has emerged as a vital paradigm for transferring the capabilities of high-capacity models to efficient ``student'' counterparts, addressing critical challenges in computational cost, deployment constraints, and privacy-sensitive settings.
Although KD is widely used in practice, it is often viewed primarily as an engineering technique, with a unified statistical perspective remaining less developed. This review bridges that gap by presenting a unified Bayesian formulation of KD that formulates teacher predictions as prior information.
This provides a principled interpretation of how teacher information is incorporated into student learning and establishes a rigorous connection to uncertainty quantification.
We demonstrate how this foundational lens reconciles classical distillation with modern extensions in generative and foundation-model systems, showing that contemporary developments remain rooted in these same statistical principles. By synthesizing theory with emerging methodologies and diverse applications, this review provides a conceptual roadmap and identifies critical open problems for the future of the field.
\end{abstract}

\end{frontmatter}

%% Keywords under the abstract (the aos layout would otherwise place them in a first-page footnote)
\begin{center}
\begin{minipage}{23pc}
\footnotesize\noindent\hspace*{16pt}{\itshape Keywords and phrases.} Bayesian inference, knowledge distillation,
large language models, model compression, uncertainty quantification.
\end{minipage}
\end{center}

\section{Introduction}

Knowledge distillation (KD) \citep{hinton2015distilling} has emerged as a central methodology for transferring predictive behavior of high-capacity ``teacher" models to more efficient ``student" models. While originally developed for neural network compression, KD now permeates modern machine learning, spanning computer vision, natural language processing, scientific modeling, and foundation-model deployment \citep{hahn2019self,liu2019structured,chen2023disentangle,guo2025deepseek,li2025frequency,xiao2025heterogeneous}. Its practical appeal lies in its ability to mitigate the prohibitive costs --- computational, financial, and environmental --- associated with deploying state-of-the-art models \citep{wu2025inference,singh2025survey}. By distilling the utility of the teacher into a compact student, KD enables powerful predictive performance in resource-constrained environments.

\begin{figure}[t]
  \centering
  \includegraphics[width=0.9\textwidth]{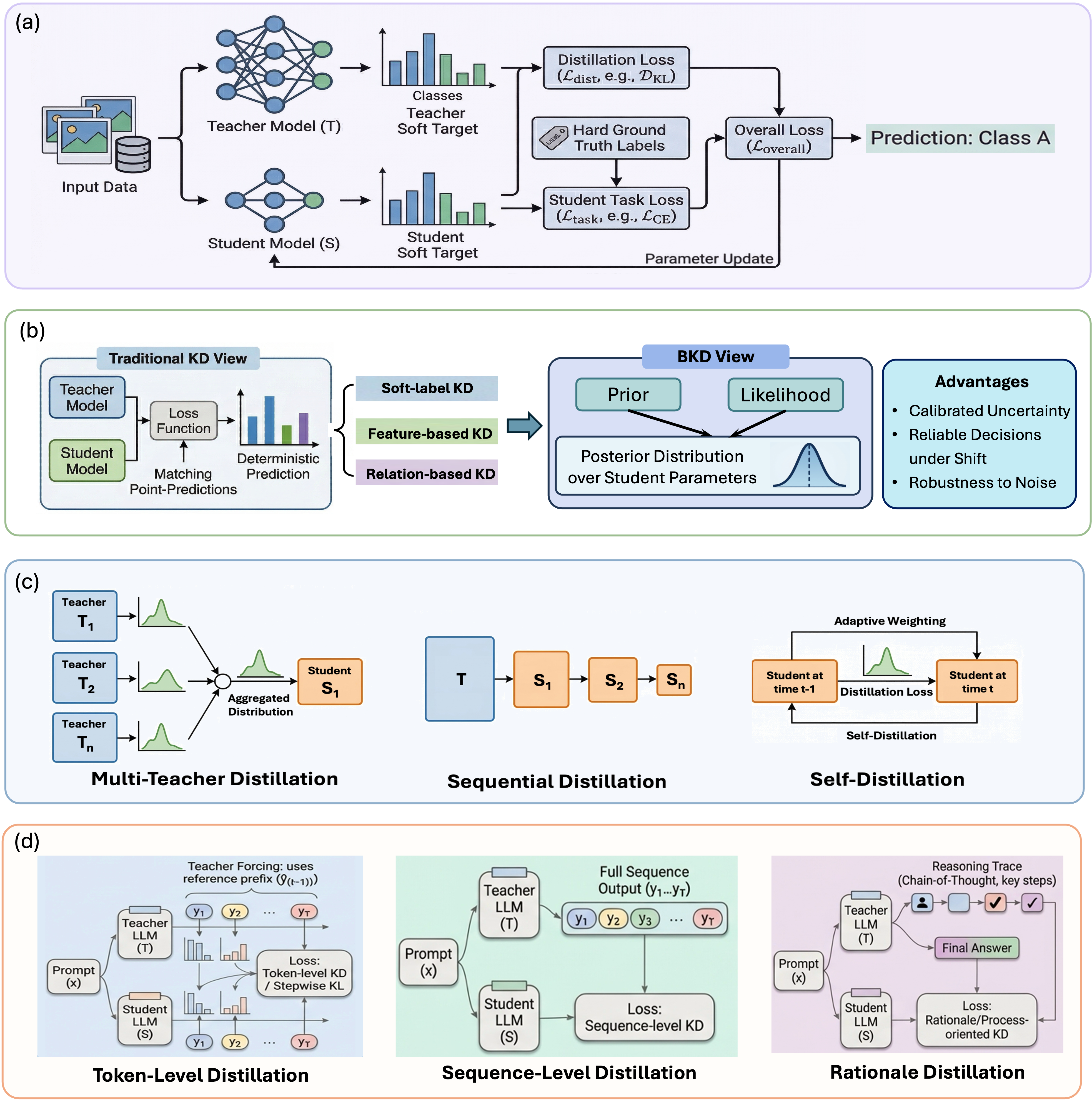}
  \vspace{-2mm}
  \caption{
  \textbf{Overview of knowledge distillation, Bayesian knowledge distillation, and representative KD methods.} (a) In classical KD, the student learns from both hard labels and teacher-derived soft targets. (b) In Bayesian KD, the teacher’s predictive distribution is incorporated as prior information and combined with the likelihood from observed labels, yielding a posterior over student parameters and associated predictive uncertainty. (c) Representative KD methods, including multi-teacher, sequential, and self-distillation. (d) Representative KD methods for LLMs, including token-level, sequence-level, and rationale distillation.
  }
  \label{fig:all}
\end{figure}

At a high level, KD trains a student model by transferring supervisory information from a teacher model, as illustrated in Figure \ref{fig:all}(a). In the classical formulation, the teacher provides a predictive distribution over labels (soft labels), which the student then treats as surrogate targets, regularizing its own predictions toward the teacher's distribution \citep{hinton2015distilling}. 
Subsequent work has expanded the object of transfer well beyond soft labels to include intermediate representations, attention patterns, geometric relations among examples, ensemble predictions, and, more recently, reasoning traces and sequence-level behavior in large language models (LLMs) \citep{romero2015fitnets,jiao2019tinybert,park2019relational,kim2016sequence,hsieh2023distilling,gu2023minillm}. As a result, KD is no longer viewed solely as a model compression technique, but more broadly as a general learning framework in which one model provides structured supervision to guide another.

%This broader view makes it important to understand KD not only as an engineering technique, but also from a statistical perspective. 
This expanded scope necessitates a shift in focus: KD should be understood not merely as an algorithmic heuristic, but as a statistical problem.
Although distillation is often introduced in terms of compression and efficiency, it fundamentally concerns how teacher-derived information is transferred to the student and used in learning. From this perspective, KD invites questions about what the student is approximating, why teacher-generated targets may improve learning beyond what is possible from observed labels alone, and how such information can be incorporated and assessed. These issues become especially crucial in modern settings, where the transferred object may be a predictive distribution, an internal representation, an ensemble summary, or an autoregressive generative process, rather than a simple class label.

Existing reviews of KD provide valuable summaries of methods and applications, but they primarily emphasize algorithmic taxonomies, empirical performance, or LLM-specific techniques \citep{gou2021knowledge,wang2021knowledge,xu2024survey,moslemi2024survey,mansourian2025comprehensive,yang2025survey}. Less emphasis has been placed on a statistical perspective that connects classical KD, theoretical understanding, and modern generative-model extensions. This perspective is increasingly important as KD expands beyond its original role in model compression and raises broader questions about how teacher-derived information should be extracted, interpreted, and transferred across different settings.

This review revisits KD through a statistical lens to provide a clearer conceptual foundation. We first outline the basic framework and classical formulations before delving into core statistical concepts, specifically early theoretical insights, a unifying Bayesian interpretation, and the challenge of imperfect teacher signals. Next, we map these principles to major extensions like multi-teacher and sequential distillation, followed by modern adaptations for LLMs. Finally, we discuss statistical evaluation and highlight real-world applications in bioinformatics, education, and foundation model deployment. Ultimately, our objective is to illustrate how statistical thinking improves the interpretation, design, and evaluation of distillation methods.

%-----taxonomy ------

%-----------
%-----------

%=========================
\section{Knowledge Distillation}\label{sec:KD}

Knowledge distillation (KD) concerns the transfer of information from a teacher model to a typically smaller or otherwise constrained student model. At a broad level, the teacher provides, for each input, a target object that conveys information about the prediction task, and the student is trained to approximate this target. The precise form of this target varies across KD methods and may involve predictive distributions, intermediate representations, relational structure, or other teacher-induced summaries.
In this section, we describe the basic framework of KD and the principal methodological forms it takes in practice.

\subsection{Problem Setup}\label{sec:problem_setup}

Let \((X,Y)\) be a random pair taking values in \(\mathcal X \times \mathcal Y\), where \(\mathcal X\) denotes the input space and \(\mathcal Y\) denotes the output space. We assume that \((X,Y)\) is jointly distributed according to an unknown population distribution \(P\), and observe training data \(D_n=\{(x_i,y_i)\}_{i=1}^n\) as independent realizations from \(P\). The goal is to construct a student predictor that performs well under \(P\) while leveraging information provided by a teacher model.

We denote the teacher by $T$ and the student by $S_\theta$, where $\theta\in\Theta$ indexes the student model class. At a broad level, KD trains the student using not only the observed labels but also additional teacher-derived information. Depending on the method, this transferred object may take the form of predictive distributions, intermediate representations, relational structure, or, in generative settings, sequence-level behavior.

Unless otherwise stated, we use multi-class classification as the running example, with $\mathcal{Y}=\{1,\dots,K\}$. For an observed label $y_i\in\mathcal{Y}$, standard supervised learning uses only the class label itself, whereas KD augments this signal with additional supervision derived from the teacher. We begin with the classical setting, where the transferred object is the teacher's predictive distribution.

%------------------------
\subsection{Classical Method: Matching Predictive Distributions}

The classical formulation of KD was introduced by \citet{hinton2015distilling}. For a fixed input $x\in\mathcal{X}$, the teacher provides a predictive probability vector
\begin{equation}
p_T(x)=\bigl(p_T(1\mid x),\dots,p_T(K\mid x)\bigr)^\top \in [0,1]^K,
\qquad
\sum_{k=1}^K p_T(k\mid x)=1,
\end{equation}
where $p_T(k\mid x)$ denotes the teacher's estimated conditional probability that \(Y=k\) given \(X=x\). The student similarly defines
\begin{equation}
p_S(x;\theta)=\bigl(p_S(1\mid x;\theta),\dots,p_S(K\mid x;\theta)\bigr)^\top \in [0,1]^K,
\end{equation}
where $\theta\in\Theta$ denotes the student model parameters. Thus, both teacher and student are viewed as producing estimated conditional class-probability distributions on the label space.

In many models, these probabilities are obtained from a real-valued score vector
\begin{equation}
\mathbf z(x)=\bigl(z_1(x),\dots,z_K(x)\bigr)^\top \in \mathbb R^K,
\end{equation}
whose entries measure the relative support assigned to the classes before normalization. These scores are converted into probabilities by the softmax transformation, 
\begin{equation}
\mathrm{softmax}(\mathbf z)_k
=
\frac{\exp(z_k)}{\sum_{j=1}^K \exp(z_j)},
\qquad k=1,\dots,K.
\end{equation}
In KD, a temperature parameter $\tau>0$ is typically introduced in the softmax transformation to produce predictive distributions,
\begin{equation}
    p^\tau(k \mid x)=\frac{\exp \left(z_k(x) / \tau\right)}{\sum_{j=1}^K \exp \left(z_j(x) / \tau\right)}.
\end{equation}
In what follows, $p_T(x)$ and $p_S(x ; \theta)$ denote the temperature-scaled predictive distributions used in the distillation term, with the temperature parameter suppressed in the notation for simplicity.

The predictive distributions expose the rich probabilistic structure encoded by the models.
For example, if an image is labeled as ``cat'', standard supervised learning uses only that hard label, whereas a teacher may assign probabilities such as $(0.72,0.20,0.06,0.02)$ to \{cat, fox, dog, rabbit\}. This indicates not only that ``cat'' is the most likely class, but also that the image resembles a fox more than a dog or rabbit. It is this additional probabilistic structure that KD transfers.

Mathematically, for a realized training pair \((x,y)\in\mathcal X\times\mathcal Y\), the standard distillation loss combines supervision from the observed label \(y\) with a discrepancy term that encourages the student to match the teacher's softened prediction:
\begin{equation}\label{eq:hinton_KD}
\mathcal{L}_{\mathrm{KD}}(x,y;\theta)
=
\mathcal{L}_{\mathrm{CE}}\bigl(p_S(x;\theta),y\bigr)
+
\lambda
D_{\mathrm{KL}}\!\left(
p_T(x)\,\|\,p_S(x;\theta)
\right),
\end{equation}
where
\(
\mathcal{L}_{\mathrm{CE}}\bigl(p_S(x;\theta),y\bigr)
=
-\sum_{k=1}^K \mathbf{1}\{y=k\}\log p_{S}(k\mid x;\theta)
\)
is the cross-entropy loss for the observed label \(y\), with \(\mathbf{1}\{y=k\}\) denoting the indicator function, equal to one if \(y=k\) and zero otherwise. 
In addition,
\(
D_{\mathrm{KL}}(p\|q)
=
\sum_{k=1}^K p_k\log(p_k/q_k)
\)
is the Kullback--Leibler divergence between predictive distributions $p$ and $q$. 

Given training data $\mathcal D_n=\{(x_i,y_i)\}_{i=1}^n$, the student parameters $\theta$ are estimated by minimizing the empirical distillation objective
\begin{equation}
\hat\theta_{\mathrm{KD}}
=
\arg\min_{\theta\in\Theta}
\frac{1}{n}\sum_{i=1}^n \mathcal{L}_{\mathrm{KD}}(x_i,y_i;\theta).
\end{equation}
From a statistical perspective, the cross-entropy term fits the student to the observed labels, while the KL term regularizes the estimated conditional class probabilities toward the teacher distribution. The parameter $\lambda>0$ controls the strength of this teacher-guided regularization.

The effectiveness of classical KD is usually attributed to two complementary effects. First, as emphasized by \citet{hinton2015distilling}, the teacher's softened predictions reveal information that is absent from the observed class labels: the probabilities assigned to non-target classes encode how the teacher generalizes and expose similarity structure among classes. With higher temperature, these targets are also smoother and higher-entropy, providing a richer supervisory signal and often yielding lower-variance gradients. Second, later work suggests that the benefit of KD is not limited to ``dark knowledge'' in the non-target classes, but also reflects a regularization effect: KD behaves in part like adaptive label smoothing, improves calibration, rescales gradients according to teacher confidence, and induces a more informative geometry in the student’s output layer \citep{tang2020understanding,yuan2020revisiting}. Thus, matching the teacher’s predictive distribution can improve the student not only by transferring class-probability information, but also by providing a smoother and better-structured training signal.

%------------------------
\subsection{Distillation of Intermediate Representations and Relational Structure}

Classical KD matches only the teacher's final output distribution, thereby ignoring much of the information contained in the teacher's internal computations. To use this additional structure, later work extends distillation to intermediate quantities such as hidden representations, attention patterns, and relations among examples \citep{romero2015fitnets,sun2019patient,jiao2019tinybert,park2019relational,tung2019similarity}. Broadly, these methods fall into two categories: representation-based distillation and relational distillation.

Representation-based methods match selected intermediate layers of the teacher and student. Let \(h_T^{(\ell)}(x)\) denote the teacher representation at layer \(\ell\), and let \(h_S^{(m(\ell))}(x;\theta)\) be the corresponding student representation, where \(m(\ell)\) maps teacher layers to student layers. When dimensions differ, an alignment map \(A_\ell\) is introduced. A generic objective is
\begin{equation}
\mathcal{L}_{\mathrm{rep}}(\theta)
=
\frac{1}{n}\sum_{i=1}^n
\sum_{\ell\in\mathcal{L}}
w_\ell\,
d_\ell\!\left(
A_\ell h_S^{(m(\ell))}(x_i;\theta),\,
h_T^{(\ell)}(x_i)
\right),
\label{eq:rep_kd_reorg}
\end{equation}
where \(w_\ell\ge 0\) are layer weights and \(d_\ell\) is a discrepancy measure such as squared distance or cosine discrepancy \citep{romero2015fitnets,sun2019patient,jiao2019tinybert}. In practice, the matched quantities may be feature maps, attention maps, or transformer layer outputs \citep{zagoruyko2017paying,wang2020minilm,wang2021minilmv2}.

Relational methods instead match how examples are arranged relative to one another in representation space, which is especially useful when teacher and student architectures differ substantially. Let \(g_T(x)\) and \(g_S(x;\theta)\) denote teacher and student representations used to define relations, and let \(\rho(\cdot,\cdot)\) be a relation map such as a distance or similarity. For a batch \(\mathcal{B}\), define
\begin{equation}
R_{T,ij}=\rho\!\left(g_T(x_i),g_T(x_j)\right),
\qquad
R_{S,ij}(\theta)=\rho\!\left(g_S(x_i;\theta),g_S(x_j;\theta)\right).
\end{equation}
A generic relational loss is
\begin{equation}
\mathcal{L}_{\mathrm{rel}}(\theta)
=
\frac{1}{|\mathcal{B}|(|\mathcal{B}|-1)}
\sum_{\substack{x_i,x_j\in\mathcal{B}\\ i\neq j}}
\delta\!\left(
R_{T,ij},\,
R_{S,ij}(\theta)
\right),
\label{eq:rel_kd_reorg}
\end{equation}
where \(\delta\) compares teacher and student relations \citep{park2019relational,tung2019similarity}. Unlike direct feature matching, this approach preserves relative geometry without requiring coordinate-wise alignment.

In practice, these internal losses are usually combined with the output-level KD objective in Equation~\ref{eq:hinton_KD},
\begin{equation}
\mathcal{L}(\theta)
=
\mathcal{L}_{\mathrm{KD}}(\theta)
+
\lambda_{\mathrm{rep}}\,\mathcal{L}_{\mathrm{rep}}(\theta)
+
\lambda_{\mathrm{rel}}\,\mathcal{L}_{\mathrm{rel}}(\theta),
\label{eq:hybrid_kd_reorg}
\end{equation}
where \(\lambda_{\mathrm{rep}},\lambda_{\mathrm{rel}}\ge 0\) control the strength of the additional supervision.

From a statistical perspective, these methods enrich distillation by constraining not only the final predictive distribution but also selected internal structure. Their success depends on whether the matched quantities reflect transferable information rather than teacher-specific artifacts, which is why relational objectives are often more robust when the teacher and student are heterogeneous.

%=========================
\section{Statistical Perspectives on Knowledge Distillation}\label{sec:stat}

The formulations reviewed in the previous section raise broader statistical questions about KD as a learning framework. These include what object the student is estimating, why soft teacher targets can improve learning, how the reliability of teacher information should be understood, and how predictive uncertainty should be represented. Existing work has studied these questions from several perspectives, including Bayesian predictive distillation, learning-theoretic analysis, and risk-based interpretations of soft labels. In this section, we review these perspectives and then discuss a Bayesian formulation of distillation that interprets teacher information as prior knowledge, recovers the classical KD objective as a posterior-mode estimator, and naturally connects distillation with uncertainty quantification.

%----------------
\subsection{Early Statistical and Theoretical Perspectives}

Several earlier works have contributed to a statistical understanding of distillation from different perspectives. 
One line of work examines why distillation can improve learning from the viewpoint of learning theory and optimization. In a tractable setting of linear and deep linear classifiers, \citet{phuong2019towards} showed that distillation can enjoy favorable generalization behavior and identified several key factors underlying its effectiveness, including data geometry, optimization bias, and a strong monotonicity property with respect to the transfer set size. These results clarify mechanisms through which soft teacher targets can influence learning beyond hard labels. However, since the analysis is developed in a highly structured setting, it does not directly extend to a general statistical formulation of KD across broader model classes.

A second perspective framed distillation through the Bayes class-probability function. \citet{menon2021statistical} observed that the population classification risk depends on the conditional class probabilities and argued that teacher soft labels can improve learning by providing lower-variance targets than the hard class labels. Their analysis establishes a bias--variance tradeoff for teacher quality and clarifies that a teacher that distills well is not necessarily the most accurate one, but rather one whose probability estimates better approximate the underlying class probabilities. This provides a statistical explanation for the role of soft labels and the importance of calibration, but it remains primarily a risk-based view of KD rather than a unified statistical formulation of how teacher information is incorporated into the student learning problem.

A third line of work considered distillation in explicitly Bayesian predictive terms. \citet{korattikara2015bayesian} proposed \emph{Bayesian dark knowledge}, where the teacher is a Monte Carlo approximation to a Bayesian posterior predictive distribution rather than a single deterministic network. The student is trained to match this predictive distribution, allowing the uncertainty-aware behavior of the Bayesian teacher to be compressed into a single model and thereby reducing the need for costly posterior or ensemble averaging at prediction time. Related work similarly explored distilling predictive distributions from Bayesian models or ensembles into compact student networks \citep{vadera2020generalized,vadera2019assessing,malinin2019ensemble}. This line of work shows that distillation can serve not only model compression but also predictive approximation for Bayesian or ensemble-based inference, but it does not, by itself, provide a general probabilistic formulation of the standard KD objective.

Taken together, these works make clear that distillation is more than a heuristic compression trick: it can be understood through Bayesian prediction, statistical risk, and optimization dynamics. At the same time, they illuminate different aspects of the problem rather than providing a unified statistical framework. This motivates formulations that interpret the teacher signal as prior information for the student, linking classical KD to Bayesian inference and naturally enabling uncertainty quantification and related extensions.

%================================
\subsection{Bayesian Knowledge Distillation}

A more explicit and unified formulation is Bayesian knowledge distillation (BKD) \citep{fangbayesian2024}, which places distillation within a Bayesian inference framework. The key idea is to treat the student parameters as random and to interpret the teacher's predictive distribution as prior information for the student. This perspective not only reinterprets the classical KD objective but also provides a coherent statistical framework in which teacher guidance, observed labels, and predictive uncertainty are handled jointly, thereby linking classical KD to posterior inference and creating a foundation for later extensions.

% \noindent\textbf{Bayesian Framework.}
Classical KD already hints at a Bayesian structure,
\begin{equation}\label{eq:hinton_KL}
    \hat\theta_{\mathrm{KD}}
    =
    \arg\min_{\theta\in\Theta}
    \frac1n\sum_{i=1}^n
    \Bigl[
    \mathcal{L}_{\mathrm{CE}}\!\bigl(p_S(x_i;\theta),y_i\bigr)
    +
    \lambda\,D_{\mathrm{KL}}\!\bigl(p_T(x_i)||\,p_S(x_i;\theta)\bigr)
    \Bigr],
\end{equation}
where the cross-entropy term handles data fitting and the distillation term acts as a prior contribution. 
However, formalizing this requires rigorously defining a likelihood and a prior on the student.

The likelihood is standard. Let $\bfy_i$ denote the one-hot encoding of $y_i$, where $y_{ik}=1$ if $y_i=k$ and $y_{ik}=0$ otherwise. Let 
$\boldq_i(\theta):=p_S(x_i;\theta)=(q_{i1},\dots,q_{iK})^T$
be the student predictive probability vector at input $x_i$. The observed labels are modeled through the multinomial likelihood
\begin{equation}\label{eq:BKD_likelihood}
    \bfy_i\mid x_i,\theta \sim \mathrm{Multinomial}\!\bigl(1;\,\boldq_i(\theta)\bigr),
\end{equation}
which is the usual likelihood for multi-class classification.

The main challenge is prior specification. Incorporating teacher knowledge directly into a prior on $\theta$ is difficult for several reasons. First, the teacher is often effectively a black box: we observe its output probabilities, but not its internal parameters or architecture. Second, even when both models are accessible, they usually differ in size or structure, so their parameters do not admit a natural one-to-one correspondence. Third, the student parameter space is often high-dimensional, making direct prior specification on $\theta$ both conceptually and computationally difficult.

BKD addresses this difficulty by modeling teacher information in the predictive space and then transferring it to the parameter space. 
Let $\boldp_i := p_T(x_i)=(p_{i1},\dots,p_{iK})^T$
% \begin{equation}
% \boldp_i := p_T(x_i)=(p_{i1},\dots,p_{iK})^T
% \end{equation}
denote the teacher's prediction. Since both $\boldp_i$ and $\boldq_i(\theta)$ lie on the same $K$-dimensional probability simplex, it is natural to place a distribution on $\boldq_i(\theta)$ centered at $\boldp_i$. In \citet{fangbayesian2024}, this is done using a Dirichlet distribution:
\begin{equation}
f(\boldq;\boldp_i)
=
\frac{1}{B(\mathbf{1}_K+\lambda\boldp_i)}
\prod_{k=1}^K q_k^{\lambda p_{ik}},
\end{equation}
where $\mathbf{1}_K$ is the $K$-vector of ones and $B(\cdot)$ is the multivariate Beta function. This choice is natural for probability vectors and has a clear interpretation: the distribution has its mode at the teacher prediction, while $\lambda>0$ controls how strongly the teacher's belief is imposed.

This predictive-space construction induces a teacher-informed prior on $\theta$ by assigning greater prior mass to those values of $\theta$ whose predictions are more consistent with the teacher. Specifically,
\begin{equation}\label{eq:BKD_prior}
    \pi_\theta\!\bigl(\theta;\{\boldp_i\}_{i=1}^n\bigr)
    \propto
    \prod_{i=1}^n
    f\!\bigl(\boldq_i(\theta);\boldp_i\bigr).
\end{equation}
By operating through the student's predictive probabilities, this formulation gracefully sidesteps the need for architectural alignment or access to the teacher's internal states.

Combining the likelihood in Equation \ref{eq:BKD_likelihood} with this teacher-informed prior (Equation \ref{eq:BKD_prior}) yields the posterior
\begin{equation}\label{eq:BKD_post}
    p(\theta \mid \mathcal{D}_n)
    \propto
    \prod_{i=1}^n
    p(\bfy_i\mid x_i,\theta)\,
    \pi_\theta\!\bigl(\theta;\{\boldp_i\}_{i=1}^n\bigr).
\end{equation}
The negative log-posterior is
\begin{align}\label{eq:BKD_logpost}
    -\log p(\theta\mid\mathcal{D}_n)
    &=
    \sum_{i=1}^n
    \Big[
    - \sum_{k=1}^K y_{ik}\log(q_{ik})
    -
    \lambda \sum_{k=1}^K p_{ik}\log(q_{ik})
    \Big]
    + C \\
    &=
    \sum_{i=1}^n
    \Big[
    \mathcal{L}_{\mathrm{CE}}\bigl(\boldq_i,\bfy_i\bigr)
    +
    \lambda
    D_{\mathrm{KL}}\!\bigl(
    \boldp_i\,\|\,\boldq_i
    \bigr)
    \Big]
    + C,
\end{align}
where $C$ is a constant. Thus, the classical KD objective is recovered as the negative log-posterior up to an additive constant, and the usual KD estimator is exactly the maximum a posteriori (MAP) estimator under this Bayesian model.

This equivalence matters for two reasons. First, it gives classical KD a principled statistical interpretation: the cross-entropy term acts as the data likelihood, while the distillation term acts as a teacher-informed prior contribution. Second, the Bayesian formulation extends KD beyond point estimation. Instead of producing only a single fitted student, it yields a posterior distribution over student parameters, providing a natural basis for uncertainty quantification and a unified framework for later extensions, including imperfect-teacher analysis, multi-teacher aggregation, and related directions discussed in this review.

% \vspace{3pt}
% \noindent\textbf{Uncertainty Quantification.}
% Beyond providing a statistical interpretation of the classical KD objective, the Bayesian formulation yields a posterior distribution over $\theta$ and therefore naturally supports uncertainty quantification. 

Since the posterior density is analytically intractable due to the nonlinear dependence of $p_S(x;\theta)$ on $\theta$, approximate sampling methods like stochastic gradient Langevin dynamics (SGLD) are typically employed \citep{welling2011bayesian,xu2018global,liu2001monte}. A generic SGLD update takes the form
\begin{equation}
    \theta^{(j+1)}
    =
    \theta^{(j)}
    +
    \eta_j \nabla_\theta \log p\!\left(\theta^{(j)}\mid \mathcal{D}_n\right)
    +
    \sqrt{2\eta_j}\,\varepsilon^{(j)},
\end{equation}
where $\eta_j$ is a step size and $\varepsilon^{(j)}\sim N(0,I)$. 
Since the negative log-posterior in Equation~\ref{eq:BKD_logpost} is equivalent to the empirical KD objective up to an additive constant, the gradient of the log-posterior is the negative gradient of the KD objective. Thus, posterior sampling can reuse essentially the same stochastic gradient computations as standard distillation training, making BKD computationally compatible with routine large-scale neural network optimization.
% Equation~\ref{eq:BKD_logpost} implies that the gradient of the log-posterior density coincides with the gradient of the classical KD objective up to sign. Thus, posterior sampling can exploit essentially the same stochastic gradient computations used in standard distillation training, making BKD computationally compatible with routine large-scale neural network optimization. 

The resulting posterior draws $\{\theta^{(m)}\}_{m=1}^M$ thereby allow uncertainty to be summarized through quantities such as the predictive variance, entropy, deviance, and related information measures \citep{hastie1987closer,renyi1961measures}. For a new observation $\tilde x$, denote the sampled predictions as $\tilde{\boldq}^{(m)}=p_S(\tilde x;\theta^{(m)})$ $(m=1,\dots,M)$.
Predictive uncertainty can be, for example, quantified using the posterior expected deviance:
\begin{equation}
    \overline{\Delta}
    =
    E_{\tilde{\boldq}}
    \left[
    E_{\bfy\sim \mathrm{Mul}(1;\tilde{\boldq})}
    \left\{
    \operatorname{dev}(\bfy,\tilde{\boldq})
    \right\}
    \right],
\end{equation}
Here $\operatorname{dev}(\bfy, \tilde{\boldq})=-2 \sum_{k=1}^K y_k \log \tilde{q}_k$ denotes the multinomial deviance.
The outer expectation with respect to \(\tilde \boldq\) can be approximated by Monte Carlo averaging over the posterior predictive draws \(\{\tilde \boldq^{(m)}\}_{m=1}^M\).

\begin{figure}[t]
  \centering
  \includegraphics[width=0.9\textwidth]{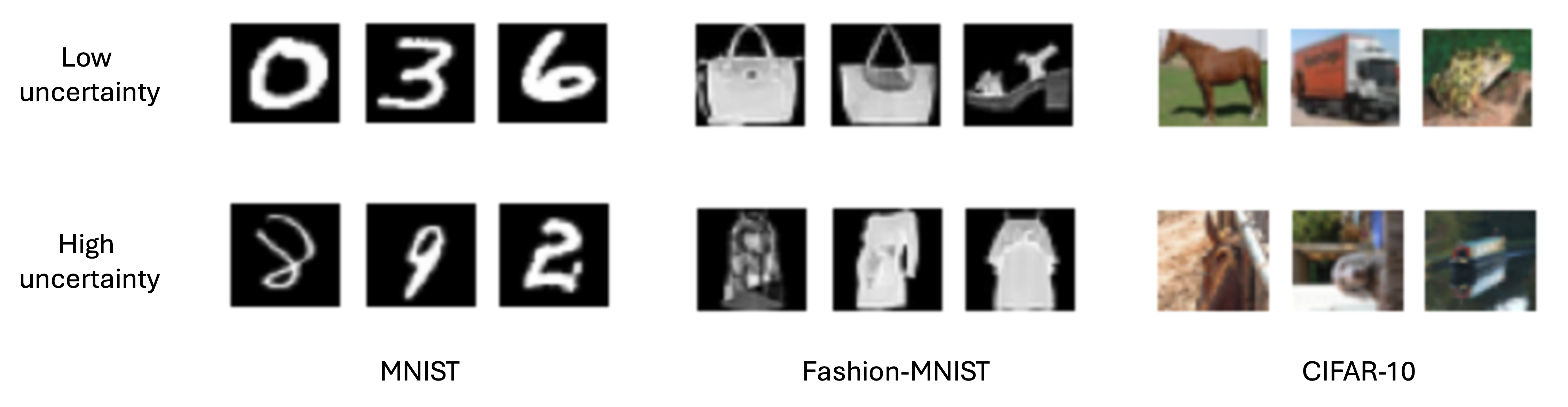}
  \vspace{-2mm}
  \caption{The top panel shows three images with the lowest mean deviance, while the bottom panel shows three images with the highest mean deviance, for (a) MNIST, (b) Fashion MNIST, and (c) CIFAR-10 datasets. Reproduced from \citet{fangbayesian2024}, licensed under CC BY 4.0.} 
  \label{fig:BKD_illu}
\end{figure}

The Bayesian formulation extends beyond point estimation by providing input-specific measures of predictive uncertainty. As illustrated in Figure \ref{fig:BKD_illu}, these measures effectively capture data quality and ambiguity: sorting examples by their posterior deviance naturally separates clear, canonical instances (top panel) from those exhibiting significant structural distortion, atypical features, or occlusion (bottom panel). This ability to identify ambiguous or potentially unreliable predictions is a substantial advantage of the Bayesian perspective, particularly in high-stakes applications where uncertainty-aware decision-making is essential \citep{gawlikowski2023survey,ai2023artificial,araujo2024road}.

%-------------------
%-------------------

% %===============================================

\subsection{Accounting for Noisy and Potentially Misspecified Teacher Models}

While KD relies on informative teacher supervision, practical teacher distributions are estimated and often noisy, biased, or miscalibrated. Under the Bayesian formulation in Section 3.2, this represents prior misspecification. This is visible in the classical KD objective:
$\mathcal{L}_{\mathrm{KD}}=\mathcal{L}_{\mathrm{CE}}\bigl(p_S(x;\theta),y\bigr)+\lambda D_{\mathrm{KL}}\!\bigl(p_T(x)\,\|\,p_S(x;\theta)\bigr).$
If $p_T(x)$ closely approximates the true conditional class probability, the distillation term provides smoother, lower variance supervision than one-hot labels alone. Conversely, a biased or poorly calibrated $p_T(x)$ pulls the student toward incorrect targets \citep{menon2021statistical}. This introduces a bias and variance tradeoff: the value of KD depends critically on the quality of the predictive distribution rather than just its raw accuracy.
Consequently, the distillation weight $\lambda$, which controls trust in the teacher, should ideally not be a global constant. When teacher reliability varies across inputs, adaptive weighting becomes necessary:
$$\mathcal{L}_{\mathrm{adapt}}=\mathcal{L}_{\mathrm{CE}}\bigl(p_S(x;\theta),y\bigr)+\lambda(x)\,D_{\mathrm{KL}}\!\bigl(p_T(x)\,\|\,p_S(x;\theta)\bigr),$$
where $\lambda(x)$ increases for trustworthy signals. Selective and confidence-aware distillation methods implement this by reducing the influence of unreliable predictions \citep{zhang2022confidence,zhang2020prime,liu2023selective,wang2024adaptive}. These methods highlight that teacher model reliability depends not only on classification accuracy, but also on the quality and calibration of the teacher's predictive distribution \citep{guo2017calibration,menon2021statistical}.

Furthermore, teacher misspecification should be distinguished from two other challenges: a structurally constrained student unable to absorb the teacher signal (approximation error) \citep{phuong2019towards,mirzadeh2020improved}, and degraded teacher reliability on synthetic or shifted inputs \citep{yin2020dreaming,ovadia2019can}. Recognizing these distinctions clarifies teacher evaluation. Since KD matches the entire predictive distribution, the best teacher is not automatically the largest model. A teacher with slightly lower classification accuracy but better calibration or more reliable uncertainty quantification may provide a more transferable signal. This explains the appeal of Bayesian and ensemble teachers, which explicitly expose predictive disagreement \citep{korattikara2015bayesian,malinin2019ensemble,fangbayesian2024}. Overall, the statistical challenge is not just to copy a stronger model, but to determine how much of the teacher signal to trust under varying conditions.

%================================
\section{Extensions of Knowledge Distillation}\label{sec:ext}

In this section, we discuss several modern extensions of KD, including multi-teacher distillation, sequential distillation, and self-distillation. These settings go beyond the basic one-teacher, one-student formulation, but they remain closely connected to the statistical perspective developed in Section \ref{sec:stat}. In each case, the central issue is how teacher-derived information should be combined, propagated, or adapted so that the student can learn effectively without inheriting unnecessary bias or instability.

%----------------------------
\subsection{Multi-Teacher Distillation}

In many practical settings, knowledge distillation must integrate information from multiple teachers. This setting is important when different teachers capture different aspects of the prediction problem, for example, because they are trained on different data sources, have different architectures, or reflect different inductive biases. In such cases, the statistical problem is not only how to transfer knowledge from a strong model to a smaller model, but also how to combine several teacher distributions into a coherent supervisory signal that the student can absorb.

Suppose we have $M$ teacher models $T_1,\ldots,T_M$, where teacher $T_m$ produces predictive distribution $p_{m}(y\mid x)$. A standard multi-teacher formulation constructs an aggregated teacher signal
\begin{equation}
p_{\mathrm{agg}}(y\mid x)
=
\sum_{m=1}^M w_m(x)\,p_{m}(y\mid x),
\qquad
w_m(x)\ge 0,
\qquad
\sum_{m=1}^M w_m(x)=1,
\end{equation}
and then trains the student to approximate $p_{\mathrm{agg}}(y\mid x)$. 
Here $w_m(x)$ is an input-dependent weight that quantifies the relative reliability of teacher $T_m$ at point $x$, with $\sum_m w_m(x)=1$. The simplest choice $w_m(x)=1/M$ recovers uniform averaging. More generally, $w_m(x)$ may vary with $x$ so that more trustworthy teachers dominate where they are reliable, for example, depending on per-teacher confidence, predictive disagreement, or domain-specific features of $x$.
Under this view, multi-teacher distillation can be interpreted as compressing an ensemble or model-averaged predictor into a single student model.

The central challenge in multi-teacher distillation is that the teachers need not be equally reliable. Uniform averaging is simple, but it can be inefficient when teacher quality varies across inputs. This has motivated adaptive weighting strategies, such as confidence-aware, disagreement-aware, and instance-dependent weighting rules for $w_m(x)$, that assign greater influence to teachers that are more trustworthy for a given input \citep{du2020agree,liu2020adaptive,zhang2022confidence}. Related collaborative frameworks further relax the fixed-teacher setting by allowing different branches or peers to act as mutual teachers during training \citep{wu2021peer}. However, most existing methods still combine teacher outputs only at the level of the training target or loss, so the aggregation rule remains largely heuristic and the resulting student is still a point estimate. Consequently, it remains difficult to represent heterogeneous teacher knowledge within a unified statistical framework, to handle input-dependent teacher relevance in a principled way, and to quantify uncertainty in the distilled student.

To tackle these challenges, a significant recent development is Bayesian multi-teacher distillation \citep{fang2026mtbkd}. Instead of using multiple teachers only to form a soft target, this framework incorporates them directly into the probabilistic specification of the student through a teacher-informed prior. A representative formulation defines a teacher-informed mixture prior of the form
\begin{equation}
\pi(\theta \mid \{p_{m}\}_{m=1}^M)
\propto
\prod_{i=1}^n
\sum_{m=1}^M
w_{m}(x_i)\,
f\!\left(p_S(\cdot\mid x_i;\theta)\mid p_{m}(\cdot\mid x_i)\right),
\end{equation}
where $f(\cdot\mid p_{m})$ is centered around the $m$-th teacher prediction and $w_{m}(x_i)$ represents the contribution of teacher $m$ for input $x_i$. This directly addresses the limitations above: it provides a coherent way to encode heterogeneous teacher knowledge, allows teacher contributions to vary across inputs within a single inferential framework, and naturally yields posterior uncertainty for the student. These features are especially valuable in applications where both predictive accuracy and uncertainty quantification matter.

%----------------------------
\subsection{Sequential Distillation} \label{sec:sequential}

When the capacity, representation, or architectural gap between the teacher and student is too large, direct transfer by KD often fails. Sequential distillation mitigates this by decomposing a difficult global transfer into a series of easier local steps, allowing knowledge to pass through intermediate models \citep{mirzadeh2020improved}. A generic pipeline can be written as
\begin{equation}
T^{(0)} \rightarrow S^{(1)} \rightarrow S^{(2)} \rightarrow \cdots \rightarrow S^{(M)},
\end{equation}
where the output of stage $m-1$ provides the teacher signal for stage $m$. The main idea is not merely to add extra models, but to smooth the transfer path. Each stage solves an easier local distillation problem than the original teacher-to-student jump, because adjacent models in the chain are more closely matched.

Sequential distillation methods mainly differ in how they build the intermediate chain. Some use assistant models with gradually reduced capacity, while others rely on repeated refinement, successive generations, or checkpoints from the teacher’s training trajectory; more complex variants further strengthen supervision with multiple intermediate teachers \citep{mirzadeh2020improved,furlanello2018born,rezagholizadeh2022prokd}. Despite these differences, the central idea is the same: replace a difficult one-step transfer with a sequence of easier local transfers. This can improve knowledge transfer when the teacher-student gap is large, but it also raises the risk of error propagation across stages.

From a statistical perspective, this structure admits a natural hierarchical Bayesian interpretation. Let $\theta_m$ denote the parameters of the model at stage $m$, with $\theta_0$ corresponding to the initial teacher. At each stage, the previous model induces a teacher-informed prior for the current student:
\begin{equation}
\theta_m \mid \theta_{m-1} \sim \pi_m(\cdot \mid \theta_{m-1}), \qquad m=1,\dots,M.
\end{equation}
Under this view, the output of stage $m-1$ is not merely an optimization target, but a prior-like constraint that shrinks the current student toward the predictive behavior of the previous stage. The full pipeline thus functions as a recursive hierarchy of shrinkage steps. Each stage combines the information inherited from the preceding model with the empirical signal of the observed data, effectively performing recursive posterior updating along the chain of models.

\subsection{Self-distillation and Variants}

In standard KD, the student learns from a fixed external teacher. When such a teacher is unavailable, computationally prohibitive, or unnecessarily restrictive, self-distillation offers an alternative by generating a supervisory signal from the model itself \citep{furlanello2018born,zhang2019byot}. Typically, this target derives from an earlier epoch or a temporal average of the student's own parameters \citep{tarvainen2017mean,caron2021dino}. Consequently, unlike standard KD, the teacher distribution in self-distillation evolves jointly with the student.
Let $\hat\theta^{(t-1)}$ denote the student parameters from the previous training stage or epoch, and define the self-teacher target,
$$
\hat p_i^{(t-1)} = p_S(\cdot \mid x_i;\hat\theta^{(t-1)}).
$$
Let \(\mathcal{L}_{\mathrm{CE}}(\theta)\) denote the cross-entropy loss with respect to the observed labels. The self-distillation objective can be written as,
\[
\hat\theta^{(t)}
=
\arg\min_{\theta}
\left[
\mathcal{L}_{\mathrm{CE}}(\theta)
+
\lambda_t \sum_{i=1}^n w_{i,t}\,
D\!\left(\hat p_i^{(t-1)},\, p_S(\cdot\mid x_i;\theta)\right)
\right],
\quad
w_{i,t}\ge 0,\quad \sum_{i=1}^n w_{i,t}=1,
\]
where $\lambda_t$ controls the overall strength of distillation and the weights \(\{w_{i,t}\}_{i=1}^n\) allow sample-wise adaptation based on factors like teacher confidence or agreement with observed labels.

Many self-distillation variants can be understood as modifications of this basic framework, designed to address the fact that the reliability of the self-teacher fluctuates over time and across samples. For instance, online distillation stabilizes the target using an exponential moving average of parameters, yielding a smoother supervisory signal \citep{tarvainen2017mean,caron2021dino}. Confidence-aware and adaptive methods dynamically adjust $w_{i,t}$ and $\lambda_t$ to down-weight unstable predictions early in training, or to provide softer targets for ambiguous examples while encouraging stricter imitation for easy ones \citep{zhang2022confidence,wei2024dtkd}.

Self-distillation has a natural empirical Bayes interpretation. From Section~\ref{sec:stat}, the teacher-informed prior is characterized by fixed center and strength parameters. In self-distillation, these quantities are estimated from the data via the model’s own predictive distribution, acting as a data-dependent prior for the next stage of training. However, this iterative updating introduces distinct statistical challenges. Since the teacher signal is estimated from the data used to fit the student, it creates cyclic dependence. While this data-dependent prior can improve regularization, it risks recycling early mistakes and reinforcing overconfident predictions if the evolving self-teacher is poorly calibrated \citep{yun2020cskd}.

A related but distinct direction further relaxes the conventional teacher-student hierarchy by allowing information to flow from weaker models to stronger ones. Research on weak-to-strong generalization shows that stronger models can improve using supervision from significantly weaker models \citep{burns2023weak,huang2020catalytic,fang2022knowledge}. In such settings, the weaker model’s predictions, learning trajectory, uncertainty, or representations can act as a regularizing signal for the larger model. This perspective shows that distillation need not be strictly unidirectional: under appropriate conditions, weaker models may help elicit latent capabilities or improve the training of stronger models.

%====================================
\section{Distillation of LLMs}\label{sec:LLM}

Large language models (LLMs) have become a central component of modern artificial intelligence and are used in a wide range of applications, including question answering, text generation, coding assistance, and scientific reasoning \citep{vaswani2017attention,openai2023gpt4,geminiteam2024gemini15}. At a technical level, an LLM defines a conditional probability distribution over text given an input prompt $x$. In practice, $x$ is itself a sequence of tokens, but for simplicity we treat it as a single object. Given $x$, the model generates an output sequence $y_{1:t}=(y_1,\dots,y_t)$ autoregressively, one token at a time, where each $y_t$ is a basic text unit such as a word, subword, or symbol. For example, if \(x=\text{``The capital of France is''}\), one possible output is
\(y_{1:2}=(\text{``Paris''},\text{``.''})\), yielding the completed text
``The capital of France is Paris.'' More generally, generation proceeds by producing each token conditional on the prompt and all previously generated tokens, first from $p(y_1\mid x)$, then from $p(y_2\mid y_1,x)$, and so on until the full sequence is formed. This autoregressive structure enables LLMs to generate coherent multi-token outputs such as explanations, summaries, and reasoning traces.

%---------------------------------
\subsection{Shared Principles and Unique Challenges in LLM Distillation}

At its core, the distillation of LLMs relies on the same principle as classical KD: approximating a teacher’s predictive distribution to yield a more efficient student. However, while the foundational mechanism of transferring information remains intact, the target of this approximation shifts dramatically.
Unlike classical KD, which targets a predictive distribution over a small and fixed label space, an LLM teacher defines a conditional generative process over token sequences through a chain of context-dependent predictions:
$$
p_T(y_{1:T}\mid x)=\prod_{t=1}^{T} p_T(y_t\mid y_{1:t-1},x),
\qquad
p_S(y_{1:T}\mid x;\theta)=\prod_{t=1}^{T} p_S(y_t\mid y_{1:t-1},x;\theta).
$$
Consequently, the student must approximate a trajectory distribution over a vast sequence space rather than a single-step prediction \citep{gu2023minillm}.

This shift introduces statistical challenges that are largely absent in classical KD. First, local errors accumulate over time because early token mismatches alter the conditioning contexts for future predictions. Second, training typically relies on teacher-forced prefixes, whereas inference relies on student-generated prefixes, creating a covariate shift mismatch in sequential prediction \citep{bengio2015scheduled,kim2016sequence}. Third, transferring broader behaviors like reasoning requires the modeling of full sequences or intermediate traces rather than just next-token probabilities \citep{hsieh2023distilling,wang2023scott}. Thus, LLM distillation must accommodate a dependent generative process where conditioning contexts evolve alongside the predictions of the model itself.

%------------------------
\subsection{Token-Level and Sequence-Level Distillation} \label{sec: 5.2}

The autoregressive formulation naturally gives rise to two related forms of distillation: one may match teacher and student locally at each next-token prediction step, or globally at the level of complete generated sequences \citep{hinton2015distilling,kim2016sequence}. In this subsection, we focus only on the pure teacher-matching component, abstracting away from any additional supervision from the original data.

Token-level distillation matches the teacher's next-token distribution at each decoding step \citep{hinton2015distilling,agarwal2024policy}. During training, this is usually done under teacher forcing, meaning that the student conditions on a reference prefix $\tilde y_{1:t-1}$ rather than on its own previously generated tokens. The reference prefix may come from the ground-truth output or from a teacher-generated sequence. A standard teacher-matching criterion is
$
\sum_{t=1}^{T}
\mathrm{KL}\!\left(
p_T(\cdot \mid \tilde y_{1:t-1},x)
\,\big\|\,
p_S(\cdot \mid \tilde y_{1:t-1},x;\theta)
\right).
$
In practice, this quantity is averaged over training examples. The appeal of token-level KD lies in its simplicity. It decomposes a long generation problem into a sum of one-step matching problems, which makes optimization relatively stable and convenient. Its limitation is that it controls the student's behavior only under the reference context distribution induced by $\tilde y_{1:t-1}$, whereas at inference time the student must condition on its own generated prefixes. As a result, token-level KD may perform well under training contexts while still suffering from exposure bias and error accumulation during free generation \citep{bengio2015scheduled,agarwal2024policy}.

Sequence-level distillation treats the full generated output as the object of transfer \citep{kim2016sequence,lin2020autokd}. Rather than aligning teacher and student one token at a time under fixed prefixes, it seeks to align their distributions over complete sequences, using
$
\mathrm{KL}\!\left(
p_T(y_{1:T}\mid x)
\,\big\|\,
p_S(y_{1:T}\mid x;\theta)
\right).
$
This formulation is more natural for generation, since the quality of an LLM depends on the full sequence produced when local predictions are rolled out over time, not only on isolated next-token probabilities. The main challenge is that the sequence space is enormous, making the exact criterion generally intractable. A common approximation is to let the teacher decode a target sequence $\hat y_{1:T}$ for input $x$, and then train the student on that teacher-generated output \citep{kim2016sequence}; the supervision comes from the teacher's preferred generation rather than directly from the original dataset. In this sense, sequence-level KD compresses the teacher's sequence-level behavior into simpler training targets, often reducing ambiguity in the supervision signal \citep{kim2016sequence}.

The Bayesian formulation in Section~\ref{sec:stat} provides a statistically principled interpretation of these criteria. Token-level distillation can be viewed as imposing teacher-informed priors on each conditional distribution $p_S(y_t\mid y_{1:t-1},x)$ under the reference context distribution, whereas sequence-level distillation constrains the joint trajectory distribution $p_S(y_{1:T}\mid x)$. From this perspective, the distinction between token-level and sequence-level KD corresponds to different choices of likelihood factorization and prior placement in a Bayesian generative model for text sequences.

%------------------------------------
\subsection{Reasoning Distillation} \label{sec:5.3}

Beyond matching final outputs, distillation can also transfer intermediate reasoning traces. A prominent example is chain-of-thought (CoT) distillation, where the teacher output is decomposed as $y_{1:T}=(y_{1:t_r},\,y_{t_r+1:T})$, with $y_{1:t_r}$ denoting the reasoning trace and $y_{t_r+1:T}$ the final answer \citep{wei2022chain}. Under this decomposition,
$
p_T(y_{1:T}\mid x)
=
p_T(y_{1:t_r}\mid x)\,
p_T(y_{t_r+1:T}\mid y_{1:t_r},x).
$
Early work typically trained the student to imitate both parts through a joint objective,
\begin{equation}
\mathcal{L}(\theta)
=
\lambda_{\mathrm{cot}}\,\mathcal{L}_{\mathrm{cot}}(\theta)
+
\lambda_{\mathrm{ans}}\,\mathcal{L}_{\mathrm{ans}}(\theta),
\end{equation}
where $\mathcal{L}_{\mathrm{cot}}$ matches the reasoning trace and $\mathcal{L}_{\mathrm{ans}}$ matches the final answer. By decomposing a difficult end-to-end mapping into intermediate steps, this form of supervision can reduce ambiguity and improve learning in smaller models \citep{hsieh2023distilling,magister2023teaching}.

However, teacher-generated rationales are not ground-truth latent variables. They may be noisy, verbose, or stylistically idiosyncratic, so exact imitation can overemphasize surface form rather than essential reasoning structure \citep{wang2023scott,feng2024keypoint}. Recent work therefore treats reasoning traces as noisy structured supervision rather than targets to copy exactly. Some methods aggregate multiple sampled chains to reduce reliance on any single rationale and emphasize self-consistency \citep{li2023symbolic,wang2023scott}. Others weight the reasoning prefix to focus on the most informative steps,
\begin{equation}
\mathcal{L}_{\mathrm{cot}}(\theta)
=
-\sum_{t=1}^{t_r} w_t \log p_S(y_t\mid y_{1:t-1},x;\theta),
\end{equation}
where a larger $w_t$ emphasizes more salient tokens. Methods such as Keypoint-based progressive CoT distillation assign higher weights to salient reasoning tokens and schedule distillation from easier reasoning steps to the full rationale \citep{feng2024keypoint}, while process supervision evaluates intermediate steps directly instead of relying only on final answer accuracy \citep{lightman2023let}.

%=========================

\section{Evaluation Frameworks and Applications}

KD affects not only model size and inference cost, but also predictive behavior, calibration, robustness, and uncertainty. Accordingly, evaluating distilled models requires more than task-level accuracy. This section reviews practical evaluation dimensions for KD and then highlights representative applications in domains where efficiency and reliable prediction are both important.

%----------------------
\subsection{Evaluation}

A distilled student may achieve similar predictive performance to its teacher while differing substantially in probability calibration, uncertainty quality, or fidelity to the teacher’s predictive behavior. More broadly, evaluation should assess both how well the student reproduces the teacher signal and how well it performs on the downstream task.

Let $p_T(y \mid x)$ denote the teacher predictive distribution and $p_S(y \mid x; \theta)$ the student predictive distribution. From a statistical perspective, evaluation can be viewed as assessing how closely $p_S$ approximates $p_T$, and how this approximation translates into predictive decisions and uncertainty estimates.

Most metrics used in the distillation literature can be grouped into three complementary categories: distributional metrics, decision-level metrics, and uncertainty metrics. These categories capture different aspects of distillation quality.

\vspace{3pt}
The most direct way to evaluate distillation is to measure the discrepancy between the teacher and student predictive distributions. Let $D(\cdot \| \cdot)$ denote a statistical divergence. Distributional metrics typically take the form
\begin{equation}
\mathbb{E}_X\!\left[D\bigl(p_T(\cdot \mid X)\|p_S(\cdot \mid X)\bigr)\right],
\end{equation}
using divergences such as the Kullback--Leibler divergence, reverse KL divergence, or Jensen--Shannon divergence \citep{hinton2015distilling,menon2021statistical}.

These metrics quantify how faithfully the student reproduces the teacher’s predictive distribution. They are especially informative when the teacher probabilities themselves carry meaningful information, such as soft class similarity or predictive uncertainty.

\vspace{3pt}
In many applications, predictions are ultimately used to make discrete decisions. Decision-level evaluation, therefore, measures how well the student reproduces the teacher’s decisions or performs relative to the ground-truth labels.

Let $\delta(x)=\arg\max_y p(y \mid x)$ denote the decision rule induced by a predictive distribution. Typical decision-level metrics include classification accuracy, F1 score, and other task-specific risk measures \citep{gou2021knowledge}. These metrics are often the most relevant for deployment, since they summarize the quality of the student’s final decisions.

Since decision-level metrics depend mainly on the induced decision boundary rather than the full predictive distribution, a student may perform well under these metrics even when its predicted probabilities differ substantially from those of the teacher.

% \subsubsection{Uncertainty metrics}

\vspace{3pt}
A third dimension of evaluation concerns predictive uncertainty. One motivation for distillation is that strong teachers, especially ensembles and Bayesian models, may provide well-calibrated probability estimates. A desirable outcome is therefore that the student preserves not only predictive accuracy but also the quality of these uncertainty estimates.

Calibration metrics such as expected calibration error (ECE) and reliability diagrams assess whether predicted probabilities agree with empirical frequencies \citep{guo2017calibration}. Unlike distributional metrics, these quantities evaluate the statistical validity of the student’s predictive probabilities rather than their closeness to the teacher.

From a statistical perspective, uncertainty metrics ask whether the student retains the teacher’s uncertainty information in a reliable and usable form. When distillation is successful, the student should preserve not only central predictions but also their associated confidence.

\begin{table}[t]
\centering
\caption{Evaluation metrics for knowledge distillation from a statistical perspective.}
\begin{tabular}{ccc}
\hline
Evaluation type & Example metrics & Statistical interpretation \\
\hline
Distributional & KL divergence, JS divergence & Fidelity of predictive distributions \\
Decision-level & Accuracy, F1 score & Quality of induced decisions \\
Uncertainty & ECE, reliability diagram & Calibration of predictive probabilities \\
\hline
\end{tabular}
\end{table}

\subsection{Applications}

KD has enabled the deployment of complex predictive models in settings where computational efficiency, latency, privacy, or interpretability are critical. Across application domains, its statistical appeal lies in approximating high-capacity or low-variance predictors while controlling generalization error, uncertainty, and robustness.
Below, we discuss some representative application areas, highlighting concrete use cases and their statistical implications.

\subsubsection{Bioinformatics and computational biology}

Bioinformatics and computational biology provide natural application domains for KD because modern models for genomic sequences, protein structures, and multi-omics data are often costly to deploy at scale. In these settings, KD is valuable not only for compression but also for transferring predictive distributions and confidence information that are important for downstream scientific use.

\begin{figure}[t]
\centering
\includegraphics[width=1.00\textwidth]{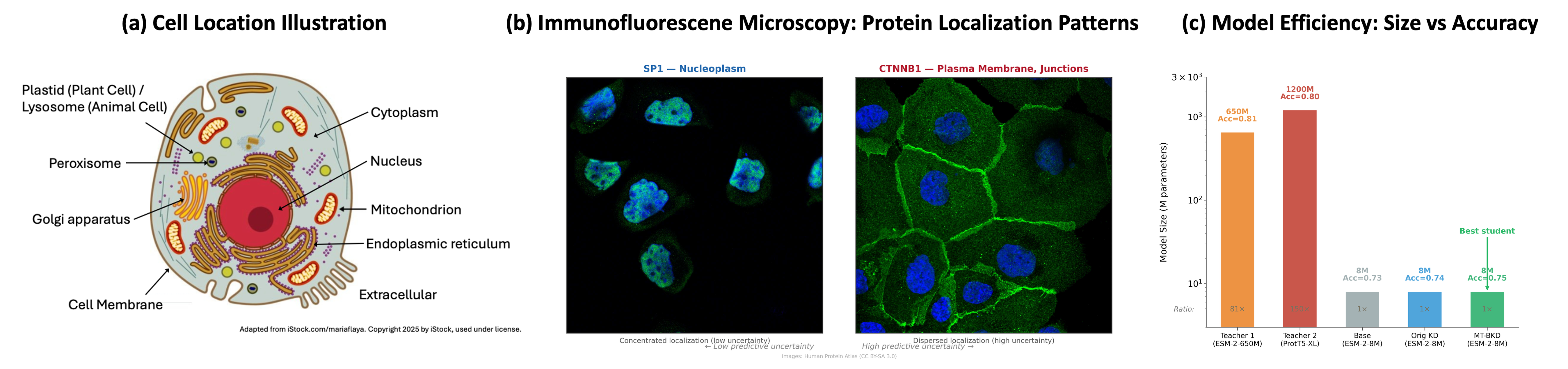}
\caption{\textbf{Multi-teacher Bayesian knowledge distillation for protein subcellular localization.}
\textbf{(a)}~Ten eukaryotic subcellular compartments in the localization task (adapted from iStock.com/mariaflaya, used under license).
\textbf{(b)}~Immunofluorescence microscopy images from the Human Protein Atlas \citep{thul2017subcellular} (CC BY-SA 3.0) illustrating two proteins for which MT-BKD produces contrasting uncertainty levels. In each image, green fluorescence marks the target protein (antibody staining) and blue marks nuclei (DAPI staining). 
\textbf{(c)}~Model size (log scale) and accuracy: the MT-BKD student (ESM-2-8M, 8\,M parameters) achieves 0.75 accuracy while being 81$\times$ and 150$\times$ smaller than the two teachers.
Data from \citet{fang2026mtbkd}.}
\label{fig:bio_kd}
\end{figure}
 
A prominent example arises in protein subcellular localization prediction, where identifying in which cellular compartment a protein resides is essential for understanding its function and for disease research \citep{hung2011protein,wang2014protein}. State-of-the-art protein language models (PLMs) such as ESM-2 \citep{lin2023evolutionary} and ProtT5 \citep{elnaggar2021prottrans} achieve strong localization accuracy but are impractical for routine laboratory use due to their large size and the fact that standard deterministic PLMs are not designed to provide calibrated uncertainty estimates. \citet{fang2026mtbkd} address both limitations with Multi-Teacher Bayesian Knowledge Distillation (MT-BKD), which distills two large PLMs, ESM-2-650M (650\,M parameters) and ProtT5-XL-half (1200\,M parameters), into a single compact student, ESM-2-8M (8\,M parameters), that is up to 150$\times$ smaller. By encoding the teachers' outputs as a mixture prior within a Bayesian framework, MT-BKD naturally yields a posterior over the student's parameters, enabling principled uncertainty quantification via posterior mean deviance and credible intervals. On the DeepLoc2 benchmark \citep{thumuluri2022deeploc}, the student retains 0.75 test accuracy (vs.\ 0.81 and 0.80 for the teachers), and its uncertainty estimates are biologically meaningful. Figure~\ref{fig:bio_kd}\,(b) illustrates two representative cases from the Human Protein Atlas \citep{thul2017subcellular}: SP1, a nucleoplasm-resident transcription factor with a strong green signal concentrated in the nucleus, receives a confident prediction, while CTNNB1 ($\beta$-catenin), whose green fluorescence spreads diffusely across the plasma membrane and cell junctions, is assigned markedly higher uncertainty, consistent with its multi-compartment localization biology (Figure~\ref{fig:bio_kd}).

KD has also been used to compress large protein language models for downstream tasks such as contact prediction and function annotation \citep{rao2021msa,lin2023evolutionary,wang2024multiPLM}. In drug discovery, distilled graph neural networks support scalable molecular property prediction, and distillation has also been combined with multi-modal embeddings for drug approval prediction \citep{sheshanarayana2025knowledge,kim2024cheMap}. In genomics and multi-omics applications, distillation can preserve diagnostic performance even when some modalities are unavailable at inference time \citep{chen2023disentangle}. Related efforts in wearable sensing and privacy-preserving medical imaging further show the value of KD for clinical and edge deployment \citep{haresamudram2024wearable,peng2025federated}.

Taken together, these applications show that, in biological and biomedical settings, distilled models often need to preserve not only predictive accuracy but also reliability and usable confidence information, since their outputs may inform downstream analysis, hypothesis generation, and real-world decision making.

%-------------------
\subsubsection{Education}\label{sec:edu}

Education is a natural fit for KD because many assessment systems face latency and hardware constraints, often running on standard laptops or tablets rather than specialized servers. For example, while large transformer-based models achieve high accuracy in automated scoring, their inference costs hinder real-time use. \citet{latif2024knowledge} addressed this by distilling a fine-tuned BERT teacher for science and math assessment into a much smaller student model. The resulting student maintains near-teacher accuracy while enabling deployment on lower-resource infrastructure, effectively converting rich teacher supervision into a deployable assessment tool.

Beyond scoring, KD compresses conversational and pedagogical models for tutoring and question answering, as seen in CourseGPT-ZH \citep{qu2024coursegpt}. Related work has similarly applied KD to domain-specific instructional models in scientific settings \citep{baladon2023retuyt,fang2026generalizable}, extending KD’s utility to interactive and domain-adapted AI.

These applications demonstrate that educational AI requires balancing predictive quality with latency, robustness, and accessibility. KD provides a principled trade-off: a modest loss in fidelity for substantial gains in reach. However, a key limitation remains: structured scoring tasks transfer more readily through KD than open-ended dialogue or reasoning. Future research should prioritize preserving calibration, explanation quality, and robustness across diverse learner populations.

\subsubsection{Large-Scale Foundation  Models}\label{sec:foundation_llm}

Foundation models, large neural networks (typically with billions of parameters) that are pretrained on broad data and subsequently adapted to many downstream tasks, have become central to modern AI.
The rise of foundation models has expanded KD's role from simple model compression to a broader mechanism for transferring frontier-scale capabilities (i.e., the largest, most capable models at the current research frontier, such as GPT-4-class and DeepSeek-R1-class models) into deployable systems \citep{hsieh2023distilling,gu2023minillm,agarwal2024policy}. 

Such models are often impractical for routine serving for two concrete reasons. First, the per-query compute and memory cost is high enough that inference must be batched on multi-GPU clusters, which translates into latency and dollar costs that are incompatible with high-throughput applications such as API serving or on-device assistants. Second, weight access and on-premise deployment are frequently restricted by licensing or privacy considerations. KD addresses both issues by propagating advanced behaviors of these large models into smaller, more deployable student models.

A prominent example is DeepSeek-R1, which shows that complex reasoning can be transferred through distillation to substantially smaller dense models. In this setting, the student is trained to reproduce the teacher’s generated outputs, including intermediate reasoning steps and multi-step solution structures, so the supervision signal is a sequence rather than a single label \citep{guo2025deepseek,hsieh2023distilling}. 
Section \ref{sec:5.3} discusses the statistical formulation of this reasoning-level supervision in more detail.
Such distilled LLMs are increasingly vital for API serving, enterprise assistants, and privacy-sensitive applications, where the objective is to optimize the trade-off between capability, latency, and inference cost \citep{gu2023minillm,zhong2024revisiting}.

This shift also redefines the nature of the transfer. While classical KD focuses on fixed label spaces (e.g., a finite set of classes in image recognition), foundation-model KD targets instruction-following, stylistic preferences, and multi-step reasoning patterns \citep{hsieh2023distilling,agarwal2024policy}. 
From a statistical standpoint, this changes the divergence being minimized from a comparison between two probability vectors over a fixed class set to a comparison between two distributions over variable-length sequences, which raises distinct estimation and evaluation challenges (see Section \ref{sec: 5.2}).
However, smaller students may still struggle with reasoning depth, robustness, or calibration in complex settings \citep{gu2023minillm,zhong2024revisiting}. Consequently, in the LLM era, KD is best understood as structured capability transfer under deployment constraints, where the goal is to retain as much ``frontier'' intelligence as the target hardware allows.

%=========================
\section{Discussion and Future Directions}

Knowledge distillation has evolved well beyond its original role as a practical compression technique. Across soft-label transfer, intermediate representation matching, multi-teacher aggregation, self-distillation, and modern LLM settings, KD is fundamentally a statistical learning problem involving estimation, regularization, and uncertainty. The central perspective of this review is that teacher-derived information should be viewed not merely as an auxiliary optimization signal, but as structured supervision that shapes the student’s learning target. Under this view, KD concerns how to combine observed data and teacher guidance in a principled and reliable way.

A recurring theme across these settings is that the main challenge in KD is not simply how to copy a larger model, but how to use teacher information appropriately. In classical classification, this appears as the problem of balancing observed labels against teacher probabilities and deciding how much trust to place in a teacher that may be noisy, biased, or miscalibrated. In richer forms of distillation, including feature-based transfer, relational distillation, multi-teacher aggregation, self-distillation, and reasoning distillation for LLMs, the same issue reappears in more complex form: the transferred object may be informative, but it is rarely exact, invariant, or uniformly reliable. From a statistical perspective, KD is therefore best understood as a family of learning problems centered on structured supervision under uncertainty, where the key questions are what is being transferred, how strongly it should influence the student, and which aspects of the teacher signal are worth preserving.

The Bayesian perspective is especially useful because it provides a coherent framework for these questions. Interpreting teacher predictions as prior-like information clarifies the role of the classical KD objective, gives a principled meaning to the distillation weight, and extends the framework beyond point estimation to uncertainty quantification. It also helps unify directions often treated separately, including imperfect-teacher analysis, multi-teacher distillation, and sequential transfer. More broadly, it shows that KD is not only about preserving predictive accuracy under resource constraints, but also about preserving reliability, calibration, and usable uncertainty in the distilled model.

This review selectively emphasizes representative developments illuminating this statistical perspective rather than exhaustively cataloging methods. For broader taxonomies or comprehensive surveys focused on methodology, readers may refer to recent surveys \citep{fang2026knowledge,mansourian2025comprehensive,yang2025survey}.

\subsection*{Summary Points}
\begin{enumerate}
    \item Knowledge distillation (KD) is both a model compression strategy and a statistical learning problem in which teacher-derived information serves as structured supervision for the student.
    
    \item Classical KD admits a natural Bayesian interpretation, where observed labels contribute a likelihood term and teacher predictions contribute prior-like information.
    
    \item The effectiveness of KD depends critically on teacher quality: calibration, uncertainty, and robustness of the teacher signal often matter more than raw teacher accuracy alone.
    
    \item Modern KD extends beyond soft-label matching to richer transfer objects, including intermediate representations, relational structure, ensemble summaries, sequential supervision, and reasoning trajectories in LLMs.
    
    \item Evaluating KD requires going beyond downstream accuracy to assess fidelity to teacher predictions, uncertainty quality, robustness, and other deployment-relevant properties.
\end{enumerate}

\subsection*{Future Issues}
\begin{enumerate}
    \item Develop principled methods for assessing teacher reliability and adaptively weighting teacher information across inputs, tasks, and stages of distillation.
    
    \item Extend probabilistic and Bayesian formulations of KD to better handle multi-teacher aggregation, sequential transfer, self-distillation, and uncertainty propagation in generative settings.
    
    \item Clarify which forms of teacher supervision, such as intermediate features, relational signals, and LLM reasoning traces, capture transferable structure rather than architecture-specific or stylistic artifacts.
    
    \item Build broader evaluation frameworks for KD that jointly assess predictive accuracy, calibration, distributional fidelity, robustness under shift, and practical deployment efficiency.
\end{enumerate}

%% ---------------------------------------------------------------------------
\begin{acks}[Acknowledgments]
During manuscript preparation and revision, the authors used generative AI tools for language editing and for generating some small figure icons (Figure~\ref{fig:all}). All such materials were reviewed and approved by the authors, who take full responsibility for the final content.

The authors are not aware of any affiliations, memberships, funding, or financial holdings that might be perceived as affecting the objectivity of this review.
\end{acks}

\begin{funding}
This work was partially supported by the U.S. National Science Foundation grants DMS-2124493, DMS-2311297, DMS-2319279, and DMS-2318809, and by the National Institutes of Health grant R01GM152814.
\end{funding}

%% ---------------------------------------------------------------------------
%% References (author-year IMS style)
\bibliographystyle{spr-ims-nameyear}
\bibliography{ref}

\end{document}